# Exploration of lane-changing duration for heavy vehicles and passenger cars: a survival analysis approach

Yang Li, Linbo Li*, Daiheng Ni

**Yang Li**
Ph.D., Candidate
Key Laboratory of Road and Traffic Engineering of Ministry of Education,
Tongji University, China,
4800 Cao'an Road, Shanghai, 201804,
E-mail: cc960719@tongji.edu.cn

**Linbo Li** (corresponding author)
Ph.D., Associate Professor
Key Laboratory of Road and Traffic Engineering of Ministry of Education,
Tongji University, China,
4800 Cao'an Road, Shanghai, 201804
E-mail: llinbo@tongji.edu.cn

**Daiheng Ni**
Ph.D., Professor
Civil and Environmental Engineering,
University of Massachusetts Amherst, Massachusetts 01003, USA
E-mail: ni@engin.umass.edu

**Abstract**
Lane-changing (LC) behavior describes the lateral movement of the vehicle from the current-lane to the target-lane while proceeding forward. Among the many research directions, LC duration (LCD) measures the total time it takes for a vehicle to travel from the current lane to the target lane, which is an indispensable indicator to characterize the LC behavior. Although existing research has made some achievements, less attention has been paid to the research of heavy vehicles' LCD. Therefore, this paper aims to further explore the LCD between heavy vehicles and passenger cars. LC trajectories are extracted from the newly-released HighD dataset, which contains of 16.5 hours of measurement and over 11,000 vehicles. The survival function of LCD has been estimated, and the characteristic has been analyzed. Thereafter, the Accelerated Failure Time model is introduced to explore the influencing factors. Results demonstrate that the MST value of passenger cars and heavy vehicles is about 5.51s and 6.08s. The heavy vehicles would maintain a longer time-headway and distance-headway with preceding vehicle when performing LC. Nevertheless, these two factors do not significantly affect the LCD of heavy vehicles. Finally, the results and the modeling implications have been discussed. We hope this paper could contribute to our further understanding of the LC behaviors for heavy vehicles and passenger cars.

# 1 INTRODUCTION

Along with car following (CF) maneuver, LC maneuver is also very common in real traffic environment. Since more interaction with surrounding vehicles are simultaneously involved, the impact of LC maneuver on traffic flow is more pronounced than that of CF maneuver, which is more likely to cause traffic accidents. Numerous researches indicate that the number of traffic accidents caused by LC maneuver maintains at a high level. Therefore, it is imperative for us to have a deep comprehension of the mechanism of LC maneuver. Over the past decades, tremendous efforts have been made on researching the decision-making process of LC [1, 2], LC trajectory planning and tracking [3-6], the impacts of LC on surroundings [7], the LC duration of the subject vehicle [8-10], etc. This paper focus on the research of LC duration, which measures the total time spent of the vehicle in the execution of LC (as shown in Figure 1). For the convenience of subsequent research, we denote LC duration as LCD.

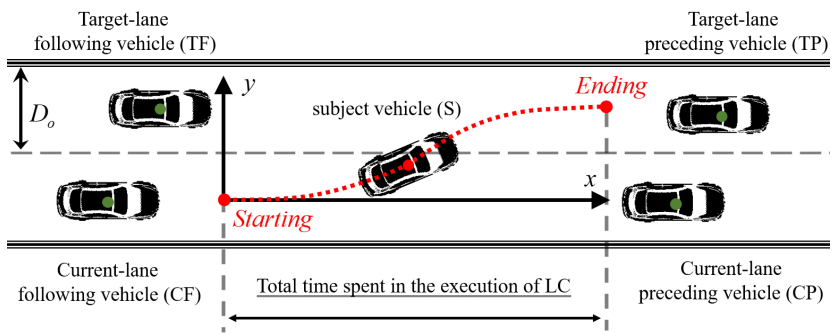

**Figure 1 The schematic diagram of LC behavior (dot curve represents the LC trajectory from the starting point to the ending point)**

LCD is a very important indicator for us to comprehend the LC behavior of vehicles. Up to now, numerous studies have explored the overall distribution and the influencing factors of LCD. Toledo and Zohar [10] employed the multiple linear regression model to analyze the LCD of the NGSIM dataset. Results demonstrated that traffic density, by the direction of the change, and by other vehicles around the subject vehicle may affect the LCD. Wu, Zhang [11] adopted the semi-parametric proportional hazard-based model to analyze the mandatory LCD data, which is collected from an unmanned aerial vehicle in a freeway maintenance construction area. Results demonstrated that there is no significant evidence showing that different vehicle types have an effect on LCD, but there is a significant difference in LCD during different time periods. Moridpour, Rose [12] investigated the effect of surrounding traffic characteristics on the LC behavior between the passenger cars and heavy vehicles. Results suggested a substantial difference of LC behaviors exists between the passenger cars and heavy vehicles. Aghabayk, Moridpour [13] compared the LC maneuvers between the heavy vehicles and passenger cars on arterial roads and freeways. Results indicated that the type and the size of vehicles influence the LC maneuver, in particular on arterial road. The LC behavior of heavy vehicles has not received appropriate attention.

Vlahogianni [14] utilized three AFT (Accelerated Failure Time) models to analyze the overtaking duration in two-lane highways. Results demonstrate that the Loglogistic distribution exhibits better performance than other models. The speed difference relative to the preceding vehicle, the speed of opposing traffic, the spacing from the lead and opposing traffic, and the driver's gender may influence the LCD. Wang, Li [15] found that there is no significance difference between the left-to-right LCD. They also conjectured that the duration times will reach a saturation value, when the velocity becomes even higher. Cao, Young [16] analyzed the difference of LC behaviors between heavy vehicles and passenger cars. A model framework for the execution of LC with regard to the emergency status and the impact of surrounding traffic for individual drivers is established. Yang, Wang [9] established a three-level mixed-effects linear regression model to explore the variables affecting LCD. The results are quite consistent with results in Toledo and Zohar [10]. Recently, Li, Li [8] presents a comprehensive analysis of LCD from the perspective of survival analysis. LC trajectories are extracted from the HighD dataset, which contains four surrounding vehicles' information at the same time. Both comparative univariate and regression survival analysis of LCD have been carried out.

Although existing research have achieved certain progress, the majority of them concentrate on studying the LCD of passenger cars, while less attention was paid on the heavy vehicles. Nevertheless, heavy vehicles' LC behavior is significantly different from passenger cars, and has more obvious impact on the real traffic flow [10, 16]. Such negligence may inevitably lead us to a one-sided understanding of LC behavior. Therefore, this paper takes a further step to research LCD of heavy vehicles. On the basis of our previous research [8], we also choose the perspective of survival analysis to explore the LCD between passenger cars and heavy vehicles. The reason why we select this perspective is due to its merit and popularity in mining the information behind the traffic data [17]. For the first time, the differences of the survival function and the influencing factors of LCD between these two types of vehicles have been investigated. At the same time, we first adopt the HighD dataset to conduct such kind of analysis, which is a new dataset of naturalistic vehicle trajectories [18]. Compared to the NGSIM dataset [10, 13, 15], this dataset is more suitable for a system-level validation of highly automated driving systems. More importantly, the proportion of trucks in this dataset is up 23% [18], while only 3% of the vehicles are trucks in NGSIM.

The above two points are the main contributions and innovations of this paper. The obtained findings and modeling implications may help us have a more comprehensive understanding of LC behaviors. The remainder of this paper is organized as follows. Section 2 presents the description and processing of the HighD dataset, and some preliminary analysis. Section 3 and Section 4 present the analysis on the difference of the survival function and the influencing factors of LCD. Finally, the conclusion is presented in Section 5.

## 2 DATA DESCRIPTION AND PROCESSING PROCEDURES

The HighD dataset is employed in this paper is recorded on German highways during 2017 and 2018 [18]. This dataset contains of 16.5 hours of measurement, 45,000 kilometers of total driven distance and over 11,000 vehicles. These trajectories are recorded in 4k (4096*2160) resolution from six different locations near Cologne, Germany. At the same time, the positioning error of each trajectory is typically less than ten centimeters. This dataset is more suitable for analyzing the LC behavior of heavy vehicles, since it has a share of 23% of heavy vehicles [18], while only 3% of the vehicles are trucks in NGSIM. For more details, please refer to Krajewski, Bock [18]. Figure 2 presents the brief introduction of the HighD dataset, including the bounding boxes of each other, the bird's eye view on the highway from a drone, and six different highway recording locations in this dataset.

We determine the beginning and ending point of LC mainly according to the lateral speed of the subject vehicle. It is worth noting that we cannot rely solely on the variable speed, but also need to combine acceleration and position information. This is because the speed and acceleration in the raw dataset are always vibrating around the value of zero [8]. Thereafter, we manually ensure that each LC trajectory is continuous and complete. It is worth noting that we only extract the successful LC trajectories. Figure 3 presents two examples of LC trajectory, which contains the longitudinal position, lateral position, lateral speed of five vehicles. Nevertheless, not all trajectories contain the information of four surrounding vehicles at the same time.

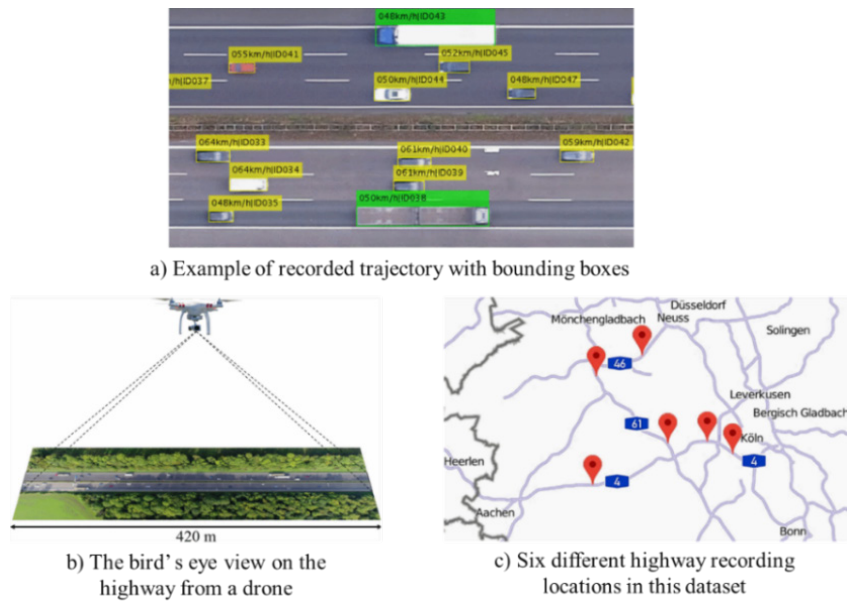

**Figure 2 The brief introduction of the HighD dataset including the bounding boxes of each vehicle, the collecting method and the recording locations**

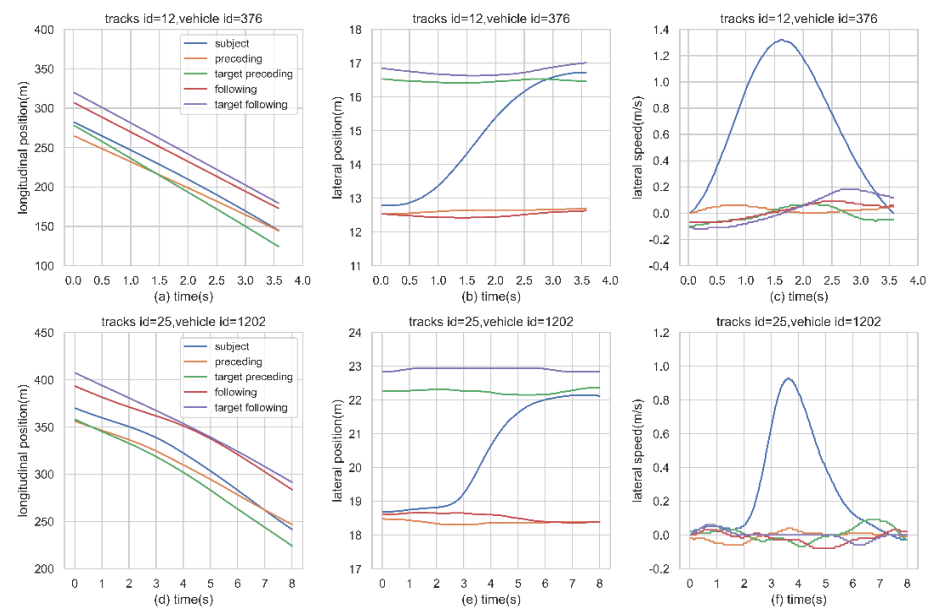

**Figure 3 Two examples of LC trajectories**

Finally, we extracted a total of 746 heavy vehicles trajectories and 7674 passenger cars trajectories. Table 1 and Figure 4 present the descriptive statistics of the LC trajectories for heavy vehicles and passenger cars. It could be found that the mean and median LCD of passenger cars is about 5.7s and 5.55s. The mean and median LCD of heavy vehicles is about 6.22s and 6.08s. This indicates that the LCD of heavy vehicles is slightly higher than of passenger cars (about 0.5s). The median value of LCD is lower than the mean value indicates that half of the vehicles pull up the mean LCD of all the vehicles.

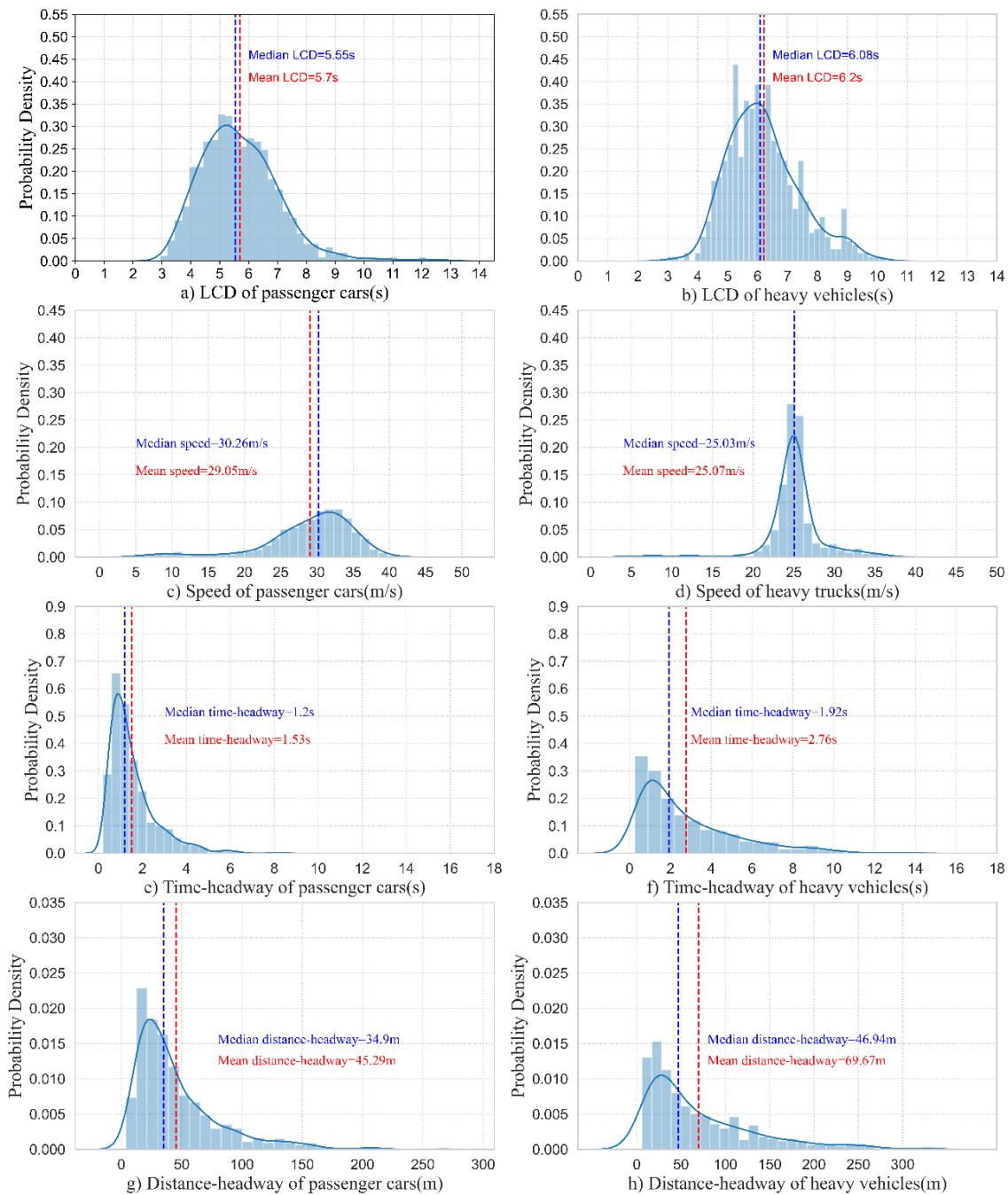

**Figure 4 Time-headway and distance-headway distribution of heavy vehicles and passenger cars**

At the same time, it could be found that the time-headway and distance-headway of heavy vehicles are all higher than that of passenger cars, while the speed of heavy vehicles is much lower than that of passenger cars. The mean time-headway of heavy vehicles and passenger cars is about 2.76s and 1.53s. The mean distance-headway of heavy vehicles and passenger cars is about 69.67m and 45.29m. The 25%th percent, 50th percent, and 75th percent of the speed of the heavy trucks are relatively close (around 25m/s). This indicates that the LC behavior of heavy vehicles is significantly different from that of passenger cars. Compared with passenger cars, the heavy vehicles would maintain a longer time-headway and distance-headway with preceding vehicle.

**Table 1 Descriptive statistics of the LCD of heavy vehicles and passenger cars**

| Descriptive statistics | | mean | std | minimum | 25th percentile | 50%th percentile | 75%th percentile | maximum |
|---|---|---|---|---|---|---|---|---|
| Passenger cars | LCD | 5.70 | 1.36 | 2.95 | 4.73 | 5.55 | 6.48 | 12.90 |
| | Speed | 29.05 | 6.34 | 4.01 | 26.45 | 30.26 | 33.11 | 47.11 |
| | Time-headway | 1.53 | 1.05 | 0.21 | 0.80 | 1.20 | 1.93 | 8.14 |
| | Distance-headway | 45.29 | 35.20 | 3.86 | 20.75 | 34.90 | 58.38 | 267.24 |
| Heavy vehicles | LCD | 6.20 | 1.20 | 3.00 | 5.33 | 6.08 | 6.88 | 10.25 |
| | Speed | 25.07 | 3.70 | 3.63 | 24.14 | 25.03 | 25.88 | 39.55 |
| | Time-headway | 2.76 | 2.36 | 0.27 | 0.99 | 1.92 | 3.90 | 13.08 |
| | Distance-headway | 69.67 | 61.33 | 6.28 | 24.83 | 46.94 | 97.72 | 334.06 |

## 3 OVERALL DISTRIBUTION DIFFERENCES

This section presents the research of the characteristic of overall survival function of LCD between heavy vehicles and passenger cars. The univariant survival model is introduced, and the obtained results are analyzed.

### 3.1 Univariant survival model

Same as our previous research, five commonly-used survival distribution functions are selected [8]. For basic concepts of survival analysis and detailed formula derivation, please refer to [8, 17, 19]. Let $T$ denotes the continuous non-negative random variable representing survival time, and we could transform $T$ into the following form:

$$Y = \log T = \alpha + \sigma W \quad (1)$$

The exponential distribution has the constant hazard value, which means the conditional probability of an event is constant over time.

$$h(t) = \lambda \quad (2)$$

Where $\lambda$ is the constant hazard. Then, we could derive $S(t) = \exp(-\lambda t)$, $f(t) = \lambda \exp(-\lambda t)$.

The Weibull distribution assumes the hazard obeys the Weibull distribution, which is capable of allowing for positive, negative, or even no duration dependence.

$$h(t) = p\lambda^p t^{p-1} \quad (3)$$

Where $p$ and $\lambda$ are the parameters that controls the shape of $h(t)$. The hazard is rising if $p > 1$, constant if $p = 1$, and declining if $p < 1$.

The Lognormal distribution assumes the $W$ has a standard normal distribution.

$$f(t) = \frac{1}{\sqrt{2\pi}\sigma t} \exp\left(-\frac{1}{2}(\frac{\log t - \mu}{\sigma})^2\right) \quad (4)$$

The $h(t)$ increases from 0 to reach a maximum and then decrease monotonically, approaching 0 as $t \to \infty$.

The Loglogistic distribution assumes the $W$ has a standard logistic distribution. The survival and hazard function are given below.

$$S(t) = \frac{1}{1+(\lambda t)^p} \quad (5)$$

$$h(t) = \frac{\lambda p (\lambda t)^{p-1}}{1+(\lambda t)^p} \quad (6)$$

Where the hazard itself monotone decreasing from $\infty$ if $p<1$, monotone decreasing from $\lambda$ if $p=1$, and similar to lognormal if $p>1$.

The Generalized-Gamma distribution assumes the $W$ has generalized extreme value distribution with parameter $k$. The density of the Generalized Gamma distribution is formulated as:

$$f(t) = \frac{\lambda p (\lambda t)^{pk-1} e^{-(\lambda t)^p}}{\Gamma(k)} \quad (7)$$

Where $p = 1/\sigma$. The Generalized Gamma includes the following interesting special cases: gamma when $p=1$, Weibull when $k=1$, Exponential when $p=k=1$, Lognormal when $k \to \infty$.

**3.2 Results and analysis**

Subgraphs (a) and (b) in Figure 5 presents the estimation of the survival function of LCD for heavy vehicles and passenger cars. Table 2 presents the corresponding AIC (Akaike Information Criterion), BIC (Bayesian Information Criterion), and MST (Median Survival Time). The AIC and BIC are both metrics of assessing model fit penalized for the number of estimated parameters. BIC penalizes model more for free parameters, and the AIC prefers a more complex over a simpler model. AIC has the danger of over fitting and BIC has the danger of under fitting. Therefore, both AIC and BIC are recommended when choosing the best parameters. It could be found that the Exponential distribution exhibits the worst performance, while the Generalized Gamma distribution outperforms than other distributions both in AIC and BIC. Therefore, we employ the Generalized Gamma distribution to conduct the subsequent analysis.

It could be found that the survival function decreased rapidly in 3s~8s, while decreased gently in 8s~12s. This indicates that most vehicles complete LC within 3s~8s. MST is defined as the time where on average 50% of the duration has expired, which indicates that each vehicle has a 50% chance of completing its LC maneuver. It could be found that the MST of heavy vehicles is 0.57s higher than that of passenger cars. From Subgraphs (c) in Figure 5, it could be found that the survival curves of these two types of vehicles are significantly different. The survival probability of heavy vehicles at each time is higher than that of passenger cars. Meanwhile, it could be found that with the increase of timeline, these two curves show a tendency to move away from each other first, and then to approach each other again. Combining with the at-risk percentage difference in subgraphs (d), we could roughly analyze that the distance between these two curves is the furthest when timeline equals to 5s. The at-risk percentage difference value reaches the highest value, roughly around 17%.

**Table 2 Fitting results of five commonly-used distribution form**

| Parametric estimator | AIC | | BIC | | MST(s) | |
|---|---|---|---|---|---|---|
| | Vehicle | Truck | Vehicle | Truck | Vehicle | Truck |
| Weibull | 3772.47 | 2157.11 | 26244.56 | 13932.99 | 5.697 | 6.26 |
| Exponential | 5795.27 | 3649.04 | 40341.62 | 23395.64 | 3.95 | 4.31 |
| Lognormal | 3514.86 | 2044.73 | 24450.76 | 13206.02 | 5.54 | 6.1 |
| Loglogistic | 3529.63 | 2057.15 | 24553.63 | 13286.34 | 5.54 | 6.07 |
| Generalized Gamma | 3511.84 | 2045.72 | 24417.83 | 13201.43 | 5.51 | 6.08 |

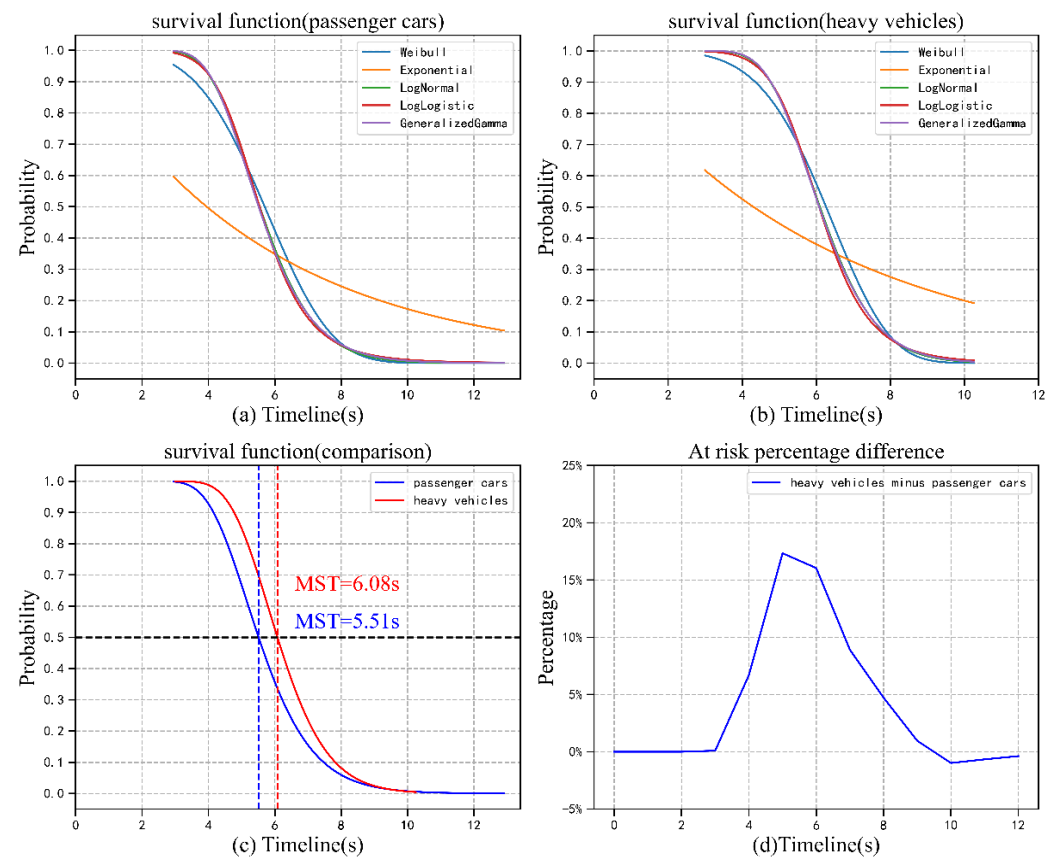

**Figure 5 The survival and cumulative hazard function of LCD using the parametric method (five commonly used distribution)**

**4 INFLUENCING FACTORS DIFFERENCE**
This section investigates the influencing factors of LCD for heavy vehicles and passenger cars. We introduce the AFT (Accelerated Failure Time) model, and analyze the regression results.

## 4.1 Accelerated Failure Time model

The AFT model directly models the survival time of the duration data, which assumes the linear relationship between the survival time and the covariates. Three AFT models (LoglogisticAFT, LognormalAFT, WeibullAFT) [8, 14] are employed to explore the influencing factors of LCD for heavy vehicles and passenger cars. The AFT model can be expressed as:

$$Y_i = x_i^T \beta + W_i \tag{8}$$

Where $Y = \log T$, $W_i$ represents the independent residuals, which is an error term with different density function. Then, we could rewrite the above Equation as:

$$T_i = T_0 \exp(x_i^T \beta) \tag{9}$$

Where $T_0 = \exp(W_i)$. When the $j-th$ dimension changes $\Delta_j$, the survival time would change $\exp(\Delta_j \beta_j)$.

The distribution assumption of the $W_i$ determine which sort of AFT model describes the distribution of the survival time $T$. The corresponding error term distributions of $W_i$ of the above three models are extreme value, normal and logistic distribution.

## 4.2 Results and analysis

Table 3 and Table 4 presents the regression results of the AFT models. It could be found that these three models achieved relatively close regression results. Compared with the LognormalAFT and the WeibullAFT model, the LoglogisticAFT model has the lowest AIC value. The corresponding MST value of passenger cars and heavy vehicles are 5.48s and 6.04s. Therefore, we employ the LoglogisticAFT to conduct the following analysis. From the regression results, it could be found that all these three variables would significantly affect the LCD of passenger cars (all the p-values are under 0.01). As for heavy vehicles, the p-values of distance-headway and time-headway are all higher than 0.5, while only the p-value of speed is under 0.005.

Take the time-headway coefficient of passenger cars for example, a unit increase of the time-headway would result in 17.4% ($e^{0.16}-1=1.174-1$) increase of baseline survival time $T_0$. For each additional unit increase of these three variables, sorted by the degree of impact, the baseline survival time of passenger cars would correspondingly increase or decrease by -0.6%, -0.4% and 17.4%. As for heavy vehicles, a unit increase of speed would result in the decrease of the baseline survival time by 1.6%.

Figure 6 presents the comparison between the baseline survival curve versus the survival curve when some covariates are varied over values. The baseline survival is calculated according to the original dataset. This is useful for us to understand subject's survival as we vary specific covariate. From Figure 6, it could be found that with the increase of the speed, both passenger cars and heavy vehicles will have shorter LCD. With the increase of time-headway, passenger cars are more likely to have a longer LCD, which is opposite to the change direction of distance-headway.

**Table 3 Regression results of LCD of passenger cars**

|  | LoglogisticAFT (AIC=3356.91, MST=5.48s) | | | LognormalAFT (AIC=3389.19, MST=5.46s) | | | WeibullAFT (AIC=3419.81, MST=5.64s) | | |
|---|---|---|---|---|---|---|---|---|---|
|  | coef | exp(coef) | p | coef | exp(coef) | p | coef | exp(coef) | p |
| Speed | -0.006 | 0.994 | 0.010 | -0.007 | 0.993 | 0.005 | -0.008 | 0.992 | 0.005 |
| Distance-headway | -0.004 | 0.996 | 0.005 | -0.004 | 0.996 | 0.005 | -0.006 | 0.994 | 0.005 |
| Time-headway | 0.160 | 1.174 | 0.005 | 0.140 | 1.150 | 0.005 | 0.210 | 1.234 | 0.005 |

**Table 4 Regression results of LCD of heavy vehicles**

|  | LoglogisticAFT (AIC=1905.16, MST=6.04s) | | | LognormalAFT (AIC=1911.72, MST=6.05s) | | | WeibullAFT (AIC=2005.59, MST=6.22s) | | |
|---|---|---|---|---|---|---|---|---|---|
|  | coef | exp(coef) | p | coef | exp(coef) | p | coef | exp(coef) | p |
| Speed | -0.016 | 0.984 | 0.005 | -0.013 | 0.987 | 0.005 | -0.014 | 0.986 | 0.005 |
| Distance-headway | 0.001 | 1.001 | 0.560 | 0.000 | 1.000 | 0.560 | 0.000 | 1.000 | 0.750 |
| Time-headway | -0.010 | 0.990 | 0.780 | 0.000 | 1.000 | 0.920 | 0.000 | 1.000 | 0.980 |

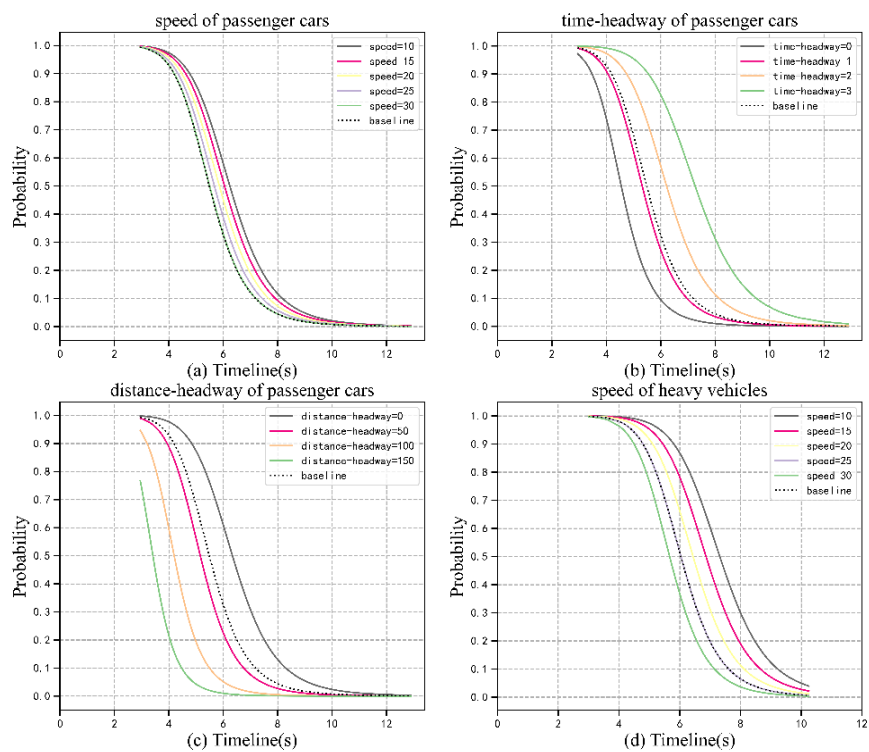

**Figure 6 Partial effects of coefficients on the LCD**

## 5 CONCLUSION

Over the past decades, tremendous efforts have been made on studying the LC behaviors of passenger cars, while neglecting the research on heavy vehicles. This paper focuses on researching the LCD for heavy vehicles and passenger cars on the basis of our previous research [8]. The univariant survival model and AFT model are employed to research the overall survival function and the influencing factors of LCD. The reason why we choose this approach is due to its capacity in formulating the relationships among the influencing factors, survival time and the outcome of each observation sample. A total of 746 LC trajectories of heavy vehicles and 7674 LC trajectories of passenger cars are extracted from the HighD dataset. The HighD dataset indicates that the average LCD of heavy vehicles and passenger cars is about 6.22s and 5.70s. Through further analysis of other variables, we found that the speed of the heavy vehicles is around 25.01m/s. The 25% percent, median, and 75% percent of the speed are relatively close to each other. Furthermore, compared with passenger cars, the heavy vehicles would maintain a longer time-headway and distance-headway with preceding vehicle as shown in Figure 4 and Table 1.

  Thereafter, the characteristics of overall survival function of LCD between heavy vehicles and passenger cars are investigated. We found that the whole survival function decreased rapidly in 3s~8s, while decreased gradually in 8s~12s. And the survival curve of heavy vehicles at each timeline is always above that of passenger cars. This also indicates that the heavy vehicles are more likely to have longer duration, and the MST difference is around 0.57s. After that, the influencing factors of LCD between heavy vehicles and passenger cars are investigated through adopting the AFT model. We found that the speed, distance-headway, and time-headway significantly affect the LCD of passenger cars, among which the most influencing factor is the time-headway. As for the, only the coefficient speed would affect the LCD. This indicates that comparted with heavy vehicles, the LC behaviors of passenger cars are more susceptible to the influence of the preceding vehicle. The LCD of a heavy vehicle is more related to its own speed. In terms of model application, the conclusions of this paper are quite consistent with our previous research [8].

  This paper presents a comprehensive analysis of LCD for heavy vehicles and passenger cars. Some novel findings have been discovered and summarized above. These findings may provide certain guidance for traffic modeling in the future. For example, these findings may guide us to reproduce the LC behavior, especially for heavy vehicles, more realistically in the traffic simulation platform. This is mainly because the execution process modeling of vehicles is often simplified or even ignored [16]. Meanwhile, these findings may provide preliminary trajectory data analysis support for the development of ADAS (Advanced Driving Assistance System). Several metrics could be adopted to assist us explore the differences in LC behavior between different regions or different time periods or different type of drivers. These metrics could be also adopted as the preliminary input variables for the LC trajectory prediction model. We hope these findings could contribute to improving our further understanding of LCD and LC behaviors.

  Undoubtedly, many aspects of this paper need further research. Given page limit, we only carried out some preliminary exploratory analysis of LC behaviors between heavy vehicles and passenger cars. One of the important research directions of this paper it to collect LC trajectory data containing more variable information, like weather, gender, roadway geometry, etc. Meanwhile, future research may try to cover a wider range of survival models, and investigate the influencing factors of unsuccessful LC events.

# 6 AUTHOR CONTRIBUTION STATEMENT

The authors confirm contribution to the paper as follows: Yang Li: Conceptualization, Data curation, Writing - original draft. Linbo Li: Methodology, Funding acquisition, Writing - original draft. Daiheng Ni: Investigation, Writing - review & editing, but with no involvement in the research grant.

# 7 ACKNOWLEDGEMENTS

This research was funded by the National Key R&D Program of China [grant numbers 2018YFE0102800].